\documentclass[default,iicol]{sn-jnl}

\jyear{2022}

\theoremstyle{thmstyleone}

\theoremstyle{thmstyletwo}

\theoremstyle{thmstylethree}

\usepackage{overpic}
\usepackage{color}
\usepackage{colortbl}
\usepackage{marvosym}

\usepackage{wrapfig}
\usepackage{booktabs}
\usepackage{multicol,multirow}

\newcommand{\figref}[1]{Fig.~\ref{#1}}
\newcommand{\tabref}[1]{Table~\ref{#1}}
\newcommand{\secref}[1]{Sec.~\ref{#1}}

\def\ie{\emph{i.e.}}
\def\eg{\emph{e.g.}}

\def\etal{{\em et al.}}

\newcommand{\zhh}[1]{{\textcolor{black}{#1}}}

\definecolor{mygray}{gray}{.9}
\definecolor{zhh}{RGB}{219, 68, 55}
\definecolor{myGreen}{RGB}{15, 157, 88}
\definecolor{myBlue}{RGB}{66, 133, 244}

\newcommand{\blu}[1]{{\textcolor{blue}{#1}}}
\newcommand{\rev}[1]{{\textcolor{red}{#1}}}

\graphicspath{{./Imgs/}{../../figure/}}

\graphicspath{{Imgs/}}

\begin{document}

\title[EFA-Net]{\textbf{Edge-aware Feature Aggregation Network for Polyp Segmentation}}

\author[1]{\fnm{Tao} \sur{Zhou}} 

\author[1]{\fnm{Yizhe} \sur{Zhang}}

\author[2]{\fnm{Geng} \sur{Chen}} 

\author[3]{\fnm{Yi} \sur{Zhou}} 

\author[1]{\fnm{Ye} \sur{Wu}}

\author[4]{\fnm{Deng-Ping} \sur{Fan}}

\affil[1]{\orgdiv{PCA Lab, Key Laboratory of Intelligent Perception and Systems for High-Dimensional Information of Ministry of Education, and the School of Computer Science and Engineering, Nanjing University of Science and Technology}, \orgaddress{\city{Nanjing}, \country{China}}}

\affil[2]{\orgdiv{School of Computer Science and Engineering}, \orgname{NPU}, \orgaddress{\city{Xi'an}, \country{China}}}

\affil[3]{\orgdiv{School of Computer Science and Engineering}, \orgname{Southeast University}, \orgaddress{\city{Nanjing}, \country{China}}}

\affil[4]{\orgdiv{Computer Vision Lab}, \orgname{ETH Zürich}, \orgaddress{\city{Zürich}, \country{Switzerland}}}



\abstract{
Precise polyp segmentation is vital for the early diagnosis and prevention of colorectal cancer (CRC) in clinical practice. However, due to scale variation and blurry polyp boundaries, it is still a challenging task to achieve satisfactory segmentation performance with different scales and shapes. In this study, we present a novel Edge-aware Feature Aggregation Network (EFA-Net) for polyp segmentation, which can fully make use of cross-level and multi-scale features to enhance the performance of polyp segmentation. Specifically, we first present an Edge-aware Guidance Module (EGM) to combine the low-level features with the high-level features to learn an edge-enhanced feature, which is incorporated into each decoder unit using a layer-by-layer strategy. Besides, a Scale-aware Convolution Module (SCM) is proposed to learn scale-aware features by using dilated convolutions with different ratios, in order to effectively deal with scale variation. Further, a Cross-level Fusion Module (CFM) is proposed to effectively integrate the cross-level features, which can exploit the local and global contextual information. Finally, the outputs of CFMs are adaptively weighted by using the learned edge-aware feature, which are then used to produce multiple side-out segmentation maps. Experimental results on five widely adopted colonoscopy datasets show that our EFA-Net outperforms state-of-the-art polyp segmentation methods in terms of generalization and effectiveness. Our implementation code and segmentation maps will be publicly at \href{https://github.com/taozh2017/EFANet}{https://github.com/taozh2017/EFANet}.
}


\maketitle

\section{Introduction}
%

\begin{figure*}[!t]
	\begin{centering}
		\includegraphics[width=0.88\textwidth]{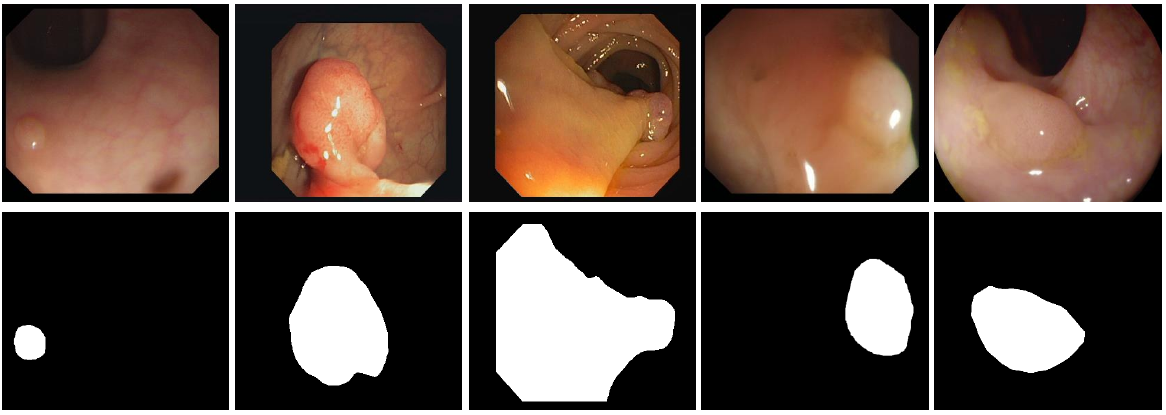}
		\caption{Illustration of different challenges in polyp segmentation, \ie, 1) The same type of polyps has a diversity of scale (see the first three columns), and 2) the blurred boundary between a polyp and its surrounding mucosa (see the fourth and fifth columns). 
  }
		\label{fig0}
	\end{centering}
\end{figure*}

\zhh{Colorectal cancer (CRC) is one of the most fatal cancer diseases worldwide~\cite{silva2014toward}, which commonly originates from adenomatous polyps. This subtype of polyps is known for their greater propensity to transform into malignant tumors if not promptly and appropriately treated}. Colonoscopy serves as a gold standard technique that can provide information on the location and appearance of polyps. However, colonoscopy is highly operator-dependent, which often misses the detection of polyps. Moreover, manual location and segmentation of polyps are subjective and time-consuming. Therefore, developing automatic polyp segmentation methods is effective in providing accurate locations of polyps for clinicians. 


Currently, U-Net~\cite{ronneberger2015u} based on an encoder-decoder architecture has turned to the prevailing framework and is widely applied to medical image segmentation tasks. Inspired by the U-Net structure, two extended models (\ie, U-Net++~\cite{zhou2019unet++} and ResUNet++~\cite{jha2019resunet++}) have been designed to achieve the polyp segmentation task and obtained good results. Enhanced U-Net~\cite{patel2021enhanced} exploited the multi-scale semantic context to enhance the feature quality for polyp segmentation. To capture more context information, ACSNet~\cite{zhang2020adaptive} utilizes the adaptive context selection scheme to explore the global and local context, which is helpful to boost the polyp segmentation performance. CCBANet~\cite{nguyen2021ccbanet} utilizes cascading context module to capture the regional and global contexts, in which a balanced attention block is introduced to conduct an attention scheme for the polyp, boundary, and background separately. 
Due to the non-sharp boundaries between polyps and their background, it is critical to locate the boundaries to help segment polyps. In this case, several methods~\cite{murugesan2019psi,fang2019selective,fan2020pra} consider capturing boundary cues or extracting boundary-aware features to boost the segmentation performance. 

Although effectiveness has been achieved, due to variations in shape, size, and location, there is still considerable room to boost polyp segmentation. 
First, one of the major challenges is scale variation in polyp segmentation. As shown in Fig.~\ref{fig0}, the first three columns show different scales and shapes of polyps. In this case, how to fully integrate the cross-level features and capture multi-scale information is still challenging. Second, as shown in the fourth and fifth columns of Fig.~\ref{fig0}, the polyps have a similar appearance to the background, where the boundaries between the polyps and their surrounding mucosa are not sharp. Thus, how to effectively utilize the boundary information or incorporate edge-enhanced features into the decoder still deserves further exploration. 


To this end, a novel Edge-aware Feature Aggregation Network (EFA-Net) is proposed for polyp segmentation in colonoscopy images, which can learn contextual knowledge from cross-level and multi-scale features, and explicitly exploit edge semantic information to benefit the segmentation performance. Specifically, we propose an Edge-aware Guidance Module (EGM), which is used to learn the edge-aware feature and then weigh the features in the decoder. Then, a Scale-aware Convolution Module (SCM) is presented to extract the multi-scale information from a single-level feature. Moreover, we proposed a Cross-level Fusion Module (CFM) to integrate the adjacent cross-level features, such cross-level fusion can enhance the representation ability of features in different resolutions. After that, the fused features are adaptively weighted by the learned edge-aware feature, and then the weighted features are used to produce the side-out segmentation maps. 
To sum up, the main contributions are three-fold:

\begin{itemize}

	\item We present a scale-aware convolution module to exploit multi-scale information and a cross-level fusion module to integrate the adjacent features, which can learn scale-correlated features and enhance their representation ability to better deal with scale variation. 

     \item An edge-aware guidance module is presented to effectively capture the edge semantics between the polyp and background, which can yield a better segmentation performance.

	\item We conduct comparison experiments on five benchmark colonoscopy datasets, and the results demonstrate that our EFA-Net achieves superior results in comparison with eleven state-of-the-art segmentation approaches. 

\end{itemize}

The rest of this paper is organized as follows. We discuss two types of works related to our study in Sec.~\ref{sec2}. Then, we provide the details of the proposed polyp segmentation framework in Sec.~\ref{sec3}. The comparison experiments are conducted in Sec.~\ref{sec4} to validate the effectiveness of our segmentation method. Finally, we draw the conclusion of this study in Sec.~\ref{sec5}.

\section{Related Works}
\label{sec2}

In this section, we briefly review some related works in polyp segmentation, boundary-based methods, and multi-scale/level feature representation.

\subsection{Polyp Segmentation}

The goal of polyp segmentation is to accurately locate and segment polyp regions from a given colonoscopy image. Traditional polyp segmentation methods mainly extract hand-crafted
features~\cite{hwang2007polyp,bernal2012towards,wang2013part,mamonov2014automated,tajbakhsh2015automated}, \eg, shape, texture, contour, and so on. However, these traditional methods often suffer from a high miss-detection rate due to the limited-expression ability of hand-crafted features. With the advancement of deep learning techniques, various convolutional neural networks (CNNs) based methods have been developed in the field of polyp segmentation. Early fully convolutional neural network (FCN) has been applied in segmenting polyps using colonoscopy images \cite{li2017colorectal}. FCN-based methods often suffer from rough segmentation outcomes prone to border errors due to low-resolution features. 

Subsequently, Lonneberger \etal~\cite{ronneberger2015u} developed a U-shaped network (U-Net) framework, which has become a popular CNN architecture for biomedical image segmentation tasks. Based on the U-Net structure, several extended models~\cite{jha2019resunet++,zhou2019unet++,ibtehaz2020multiresunet} have been developed for polyp segmentation. However, the existing U-Net structures, such as ResU-Net~\cite{zhang2018road}, H-DenseUNet~\cite{li2018h} and Attention U-Net~\cite{oktay2018attention}, directly adopt a skip connection to fuse feature representations from the encoder to the decoder. However, simple skip connections could degrade the segmentation performance due to the gap between the features of the encoder and that of the decoder. To address this issue, Zhou \etal~\cite{zhou2019unet++} proposed UNet++ by designing a dense skip connection strategy to improve the segmentation performance. Besides, several methods~\cite{murugesan2019psi,fang2019selective,fan2020pra,zhoupolyp2023,yue2022boundary} propose to learn boundary-aware features to boost the segmentation performance. For instance, Psi-Net~\cite{murugesan2019psi} simultaneously utilizes the boundary and region information, however, the relationship between the boundary and region can not be effectively explored. SFA~\cite{fang2019selective} leveraged area and boundary constraints to design a selective feature aggregation module for polyp segmentation. PraNet~\cite{fan2020pra} effectively aggregates high-level features to exploit contextual information and adopts multiple reverse attention modules to capture the correlations between regions and boundaries for calibrating some inaccurate predictions. Zhou \etal~\cite{zhoupolyp2023} designed a boundary prediction network to learn the boundary-aware features, which are incorporated into the segmentation network to boost the segmentation performance. Besides, several polyp segmentation methods~\cite{liu2022dbmf,su2023accurate,song2022attention} fully make use of multi-scale features and conduct feature fusion to enhance the model performance. Moreover, various methods~\cite{tomar2022tganet,wei2021shallow,wu2022polypseg,yin2022duplex,lin2022bsca} utilize attention mechanisms to achieve polyp segmentation.

\begin{figure*}[!t]
	\begin{centering}
		\includegraphics[width=1.0\textwidth]{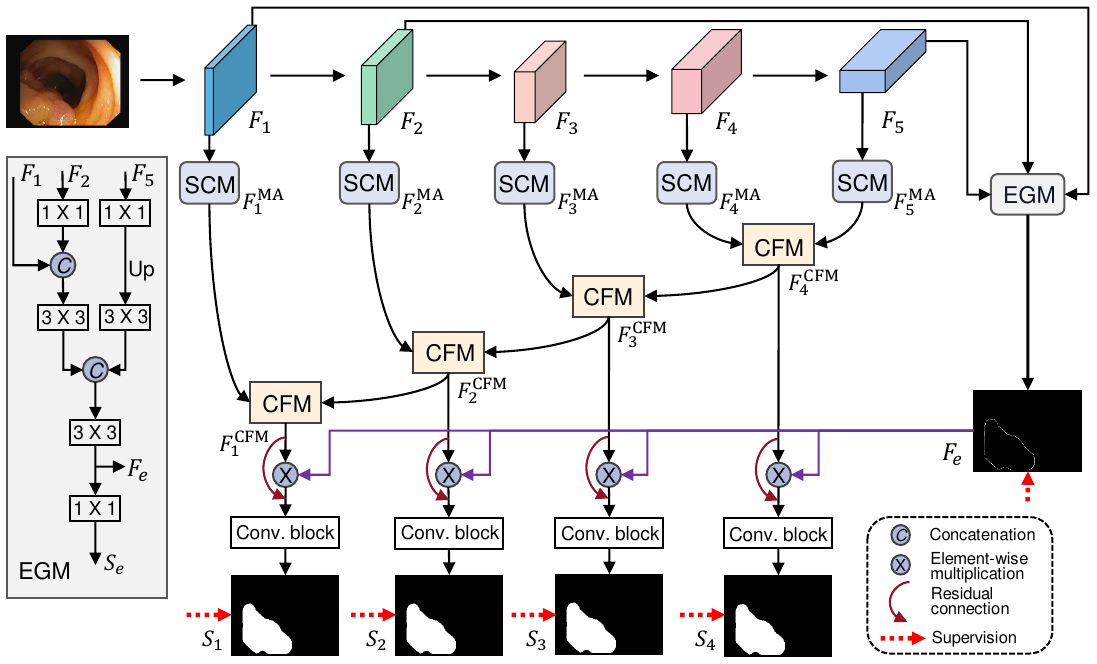}
		\caption{The architecture of the proposed EFA-Net, which consists of edge-aware guidance module, scale-aware convolution module, and cross-level fusion module..}
		\label{fig1}
	\end{centering}
\end{figure*}


\zhh{\subsection{Boundary-based Methods}}

\zhh{Boundary-based methodologies leverage edge information within the network to amplify feature representations, ultimately enhancing model performance~\cite{ding2019boundary,qin2019basnet,liu2022weakly,zhou2022feature,li2023image,nawaz2022melanoma,maskeliunas2023pareto}. For instance, Ding \etal~\cite{ding2019boundary}~innovatively introduced a boundary-aware feature propagation module. This module is designed to strengthen the similarity of local features within identical segment regions while concurrently maintaining the unique traits of features linked to diverse segments. Building upon this, Zhou \etal~\cite{zhou2022feature} formulated a boundary guidance module to learn boundary-enhanced feature representations, which can capture the local characteristics and boundary details to improve the detection performance of camouflaged objects. Beyond these applications, various research investigations have leveraged boundary/edge information to enhance medical tasks~\cite{chen2017dcan,zhang2019net,fan2020pra,ramasamy2021detection,wang2022boundary,wang2022eanet}. For example, Zhang \etal~\cite{zhang2019net} proposed a universal medical segmentation framework that utilizes edge-attention features to guide segmentation processes. Similarly, Wang \etal~\cite{wang2022eanet} introduced an edge-preserving module that incorporates the learned boundaries into the middle layer to improve the segmentation accuracy. Moreover, SFA~\cite{fang2019selective} introduces an innovative boundary-sensitive loss to mutually improve both polyp region segmentation and boundary detection. BA-Net~\cite{wang2022boundary} employs a multi-task learning approach to jointly segment object masks and identify lesion boundaries. In this model, interactive attention is designed to harness complementarity information from different tasks, thereby boosting segmentation performance.}

\subsection{Multi-scale/level Feature Representation}

Multi-scale feature representation has been widely studied in many computer vision tasks, \eg, object detection~\cite{kim2018san,cao2019high,zhoucvmj2023}, image super-resolution~\cite{li2018multi,li2020mdcn}, and semantic segmentation~\cite{he2019dynamic,gu2022multi}. Besides, two representative strategies (\ie, pyramid pooling \cite{zhao2017pyramid} and atrous spatial pyramid pooling (ASPP)~\cite{chen2017deeplab}) have been widely applied to multi-scale feature extraction. 
Currently, multi-scale or cross-level feature fusion~\cite{yue2022boundary,yang2019clci,zhao2021automatic,srivastava2021msrf,yang2023flexible} has received increasing attention in the field of medical image segmentation, which is helpful in exploiting regional context information and handling the issue of scale variations. For example, Yang \etal~\cite{yang2019clci} proposed a cross-level fusion strategy to integrate multi-scale features from different levels for chronic stroke lesion segmentation. Srivastava \etal~\cite{srivastava2021msrf} proposed a multi-scale residual fusion network for medical image segmentation, which is specially designed for medical image segmentation, in which a dual-scale dense fusion strategy is designed to exploit multi-scale information by varying receptive fields. Yue \etal~\cite{yue2022boundary} fully integrate cross-layer features to capture more context information for improving the polyp segmentation performance.  Although these methods have obtained great progress, fusing multi-scale and cross-level feature representations to capture scale variation and integrate region context information, still deserves to be further explored for boosting the segmentation performance.

\section{Methodology}\label{sec:Methods}
\label{sec3}

In this section, we first give an overview of the proposed polyp segmentation framework in Sec.~\ref{sec3-1}. Then we introduce the three key components, the edge-aware guidance module (\secref{sec3-2}), scale-aware convolution module (\secref{sec3-3}), and cross-level fusion module (\secref{sec3-4}). Further, we introduce the learning details in Sec.~\ref{sec3-5}, including model inference and loss function.  

\subsection{Overview of Architecture} 
\label{sec3-1}

\figref{fig1} presents the architecture of our EFA-Net. Specifically, the Res2Net~\cite{pami20Res2net} is adopted as the backbone to extract multi-level features for given images, denoted as $\{F_i, i=1,2,\dots,5\}$. Then, the proposed EGM combines the low-level features (\ie, $F_1$ and $F_2$) and the high-level feature (\ie, $F_5$) to learn an edge-enhanced feature, which is further incorporated into the decoder. Besides, each feature $F_i$ is fed into the proposed SCM to extract the multi-scale features. Further, we utilize the CFM to integrate the cross-level features to fully exploit the local and global contextual information. Finally, our decoder produces multiple side-out segmentation maps $\{S_i, i=1,2,3,4\}$.

\subsection{Edge-aware Guidance Module}
\label{sec3-2}

Some existing works have pointed out that edge information is able to benefit the segmentation and detection performance~\cite{fan2020inf,sun2022boundary,zhang2019net,zhao2019egnet}. 
Since low-level features preserve sufficient boundary details, existing methods~\cite{fan2020inf,zhang2019net,zhao2019egnet} usually integrate low-level features to learn edge-enhanced representations. Although low-level features contribute more information to learning edge-aware features, they could introduce non-edge coarse details. Therefore, inspired by~\cite{sun2022boundary}, we combine the two low-level features (\ie, $F_1$ and $F_2$) with the high-level feature (\ie, $F_5$) to construct our EGM. Specifically, as presented in Fig.~\ref{fig1}, $F_2$ and $F_5$ are first fed two $1\times{1}$ convolutional layers, respectively, to obtain $F_2^{'}$ and $F_5^{'}$. Then, $F_1$ and $F_2^{'}$ are cascaded and then passed through a sequential operation to obtain $F_{12}=\mathcal{C}_{3{\times}3}(\textup{Concat}(F_1,F_2^{'}))$, where $\mathcal{C}_{3\times 3}(\cdot)$ is a sequential operation that consists of a $3\times{3}$ convolutional layer, batch normalization and a \emph{ReLU} activation. Moreover, we conduct a concatenation operation on $F_{12}$ and $F_5^{'}$ to obtain the edge-enhanced feature, which can be depicted by 
\begin{equation}
\begin{aligned}
F_{e}=\mathcal{C}_{3\times 3}(\textup{Concat}(F_{12},\textup{Up}(F_5^{'}))),
\end{aligned}
\end{equation}
where $\textup{Up}(\cdot)$ denotes an upsampling operation. Finally, $F_e$ is passed through a $1\times{1}$ convolutional layer to generate an edge map ($S_e$), and the edge map is upsampled to ensure that it has the same resolution as the original image. Here we adopt the binary cross-entropy (BCE) loss (termed $\mathcal{L}_{BCE}$) to measure the difference between the produced edge map and ground-truth map ($G_e$, which is obtained by using the Sobel detection operator), and the associated loss can be defined by $\mathcal{L}_{e}=\mathcal{L}_{BCE}(S_e, G_e)$. Note that, $F_e$ provides an edge-aware feature to weigh the features in the decoder for boosting the segmentation performance.

\begin{figure}[!t]
	\begin{centering}
		\includegraphics[width=0.5\textwidth]{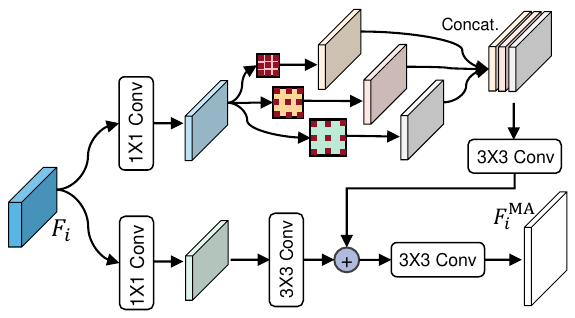}
		\caption{Architecture of the proposed scale-aware convolution module.}
		\label{fig2}
	\end{centering}
\end{figure}

\subsection{Scale-aware Convolution Module}
\label{sec3-3}

Due to the same type of polyps having different scales, it is important to learn scale--correlated feature representations to boost the model in segmenting polyps. To achieve this, we propose a Scale-aware Convolution Module (SCM) to extract scale-aware features, which can exploit different scale-correlated information to enhance the feature representations. Specifically, $F_i$ is first passed through two $1\times 1$ convolutions to obtain $F_i^1$ and $F_i^2$. Then, to learn scale-aware feature representations, we feed $F_i^1$ into three $3\times{3}$ convolutions by using different dilated rates, respectively, thus we can obtain 
\begin{equation}
\begin{aligned}
E_l=\mathcal{C}_{3\times 3}^{r_l}(F_i^{1}),
\end{aligned}
\end{equation}
where $\mathcal{C}_{3\times 3}^{r_i}$ denotes a $3\times{3}$ dilated convolution with a dilated rate of $r_i$, followed by batch normalization with a \emph{ReLU} activation. In this study, three different dilated rates, \ie, $r_i=2,4,8$, are adopted. After that, the three scale-aware features (\ie, $E_l, l=1,2,3$) are concatenated, which is passed through a $3\times{3}$ convolutional layer for adaptive aggregation to obtain the concatenated feature. At last, the concatenated feature and the original feature are fused via an addition operation, thus this process can be represented by
\begin{equation}
\begin{aligned}
F_i^{\textup{SCM}}=\mathcal{C}_{3\times 3}\big(\mathcal{C}_{3\times 3}(\textup{CAT}(E_1,E_2,E_3))\oplus\mathcal{C}_{3\times 3}(F_i^2)\big),
\end{aligned}
\end{equation}
\noindent where $\oplus$ denotes an element-wise addition operation. As a result, we obtain the scale-aware feature $F_i^{\textup{SCM}}$, which is regarded as the output of the $i$-th SCM. 

It is worth noting that our SCM can enhance multi-scale feature representations by utilizing different dilated convolutions, in which small-scale  (with small dilated rates) and large-scale (with large dilated rates) features can be aggregated to handle the challenge of scale variation. 

\subsection{Cross-level Fusion Module}
\label{sec3-4}

We propose the CFM to fuse the features from adjacent levels, which can effectively exploit cross-level contextual semantics to improve the polyp segmentation performance. Specifically, taking $F_a$ and $F_{b}$ as examples, they are first concatenated to obtain the fused feature $F_{cat}$. Then, $F_{cat}$ is passed through three sequential convolution operations, and we obtain $F_{cat}^{1}$, $F_{cat}^{2}$, and $F_{cat}^{3}$. Because local contexts can avoid small polyps being ignored, while global contexts are helpful to emphasize feature learning to locate some large polyps. Therefore, in the proposed CFM, we present to capture local contexts as well as exploit global contexts to enhance feature representations. To achieve this, in the ``Local Att." branch shown in Fig.~\ref{fig2}, we conduct two point-wise convolutions (PWC)~\cite{dai2021attentional} to learn the local attentional weights. Besides, in the ``Global Att." branch shown in Fig.~\ref{fig2}, we first carry out a global average pooling (GAP) operation and then utilize PWC operation to learn the global weights. Therefore, we can obtain the local and global attention-based weights as follows:
\begin{equation}
\left\{
\begin{aligned}
&W_{local}={\mathcal{S}}\big(\mathcal{P}_{conv2}({\mathcal{R}}(\mathcal{P}_{conv1}(F_{cat}^{1})))\big),\\
&W_{global}={\mathcal{S}}\big(\mathcal{P}_{conv2}({\mathcal{R}}(\mathcal{P}_{conv1}(\textup{GAP}(F_{cat}^{2}))))\big),
\end{aligned}
\right.
\end{equation}
where $\mathcal{P}$ denotes the point-wise convolution, and the kernel sizes of $\mathcal{P}_{conv1}$ and $\mathcal{P}_{conv2}$ are $\frac{K}{t}\times K\times 1 \times 1$ and $ K\times \frac{K}{t}\times 1 \times 1$, respectively. Here $t$ is a channel reduction ratio, and $K$ is the channel size. 
Besides, $\mathcal{S}(\cdot)$ and $\mathcal{R}(\cdot)$ indicate the \emph{Sigmoid} and \emph{ReLU} activation functions, respectively. 
Next, we conduct the element-wise multiplication between the features and the corresponding weights with residual connection, thus the enhanced features can be given by 
\begin{equation}
\left\{
\begin{aligned}
&F_{en}^1=F_{cat}^{1}\otimes{W_{local}}\oplus{F_{cat}^{1}},\\
&F_{en}^2=F_{cat}^{2}\otimes{W_{global}}\oplus{F_{cat}^{2}},
\end{aligned}
\right.
\end{equation}
where ${\otimes}$ denotes element-wise multiplication. After that, the two attention-enhanced features and the original feature $F_{cat}^{3}$ are concatenated and then fed to a sequential convolution operation to generate the final output $F_{i}^{\textup{CFM}}$, which is obtained as follows:
\begin{equation}
\begin{aligned}
F_i^{\textup{CFM}}=\mathcal{B}_{conv3{\times}3}\big(\textup{Concat}(F_{en}^1,F_{en}^2,F_{cat}^3)\big).
\end{aligned}
\end{equation}

\begin{figure}[!t]
	\begin{centering}
		\includegraphics[width=0.5\textwidth]{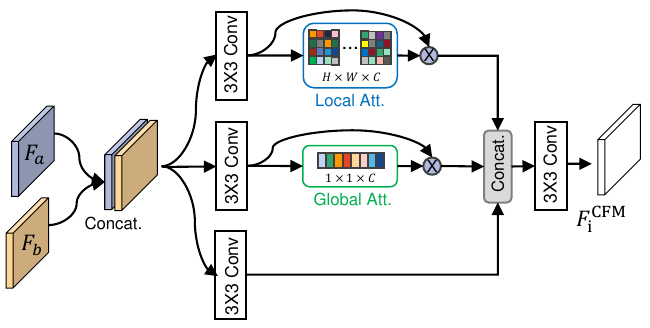}
		\caption{Architecture of the proposed cross-level fusion module.}
		\label{fig3}
	\end{centering}
\end{figure}

\subsection{Learning Details}
\label{sec3-5}

\textbf{Model Inference}: We have obtained the cross-level fused features $F_i^{\textup{CFM}} (i=1,2,3,4)$, thus it is important to integrate the edge-aware feature to provide more boundary details. To achieve this integration, the edge-aware feature $F_e$ is first passed through a \emph{Sigmoid} activation to normalize values to $[0,1]$, which can be used as an edge feature-level attention map. In order to preserve the original information from $F_i^{\textup{CFM}}$, a residual connection strategy is also adopted to combine the edge-enhanced feature with $F_i^{\textup{CFM}}$, thus obtaining the edge-aware features 
\begin{equation}
\begin{aligned}
F_i^{\textup{edge}}=F_i^{\textup{CFM}}\otimes \mathcal{S}(F_e)+F_i^{\textup{CFM}}.
\end{aligned}
\end{equation}

Then, $F_i^{\textup{edge}}$ will pass through a ``Conv. block" (as shown in Fig.~\ref{fig1}) to produce multiple side-output segmentation maps, namely $S_i~(i=1,2,3,4)$. Here ``Conv. block" contains two sequential convolutions (\ie, $\mathcal{B}_{conv3{\times}3}$) followed by a $1\times{1}$ convolution. 

\textbf{Loss Function}: Inspired by~\cite{fan2020pra,qin2019basnet,wei2019f3net}, our segmentation loss is defined by $\mathcal{L}_{seg} =\mathcal{L}_{BCE}^w+\mathcal{L}_{IoU}^w$, 
where $\mathcal{L}_{BCE}^w$ and $\mathcal{L}_{IoU}^w$ denote the weighted BCE loss and  weighted intersection over union (IoU) loss, respectively. Finally, the overall loss function is given by 
\begin{equation}
\begin{aligned}
\mathcal{L}_{total}=\sum_{i=1}^4\mathcal{L}_{seg}(S_i,G)+\beta\mathcal{L}_{e},
\end{aligned}
\end{equation}
where the hyper-parameter $\beta$ is set to $5$ in this study, and $G$ is the ground-truth map.

\begin{table*}[tp]
  \centering
  \small
  \renewcommand{\arraystretch}{1.1}
  \setlength\tabcolsep{6.2pt}
  \caption{Results comparison on the CVC-ClinicDB and Kvasir datasets. The best results are highlighted.} 
  \begin{tabular}{r||ccccc|ccccc}
  \hline
    
    \multirow{2}*{\textbf{Methods}}   
    &\multicolumn{5}{c|}{CVC-ClinicDB \cite{bernal2015wm}} 
    &\multicolumn{5}{c}{Kvasir~\cite{jha2020kvasir}}\\
    
    \cline{2-11}
    
    & mDice  & mIou  & $S_{\alpha}$  &$F_{\beta}^{w}$ & $E_{\phi}^{mean}$ 
    & mDice  & mIou  & $S_{\alpha}$  &$F_{\beta}^{w}$ & $E_{\phi}^{mean}$ \\

    \hline
  
  
    UNet~\cite{ronneberger2015u}~
    & 0.823   & 0.755   & 0.889 & 0.811   & 0.914 	
    & 0.818   & 0.746   & 0.858 & 0.794   & 0.881 \\
    
    UNet++~\cite{zhou2019unet++}~
    & 0.794   & 0.729   & 0.873 & 0.785   & 0.891 	
    & 0.821   & 0.744   & 0.862 & 0.808   & 0.887 \\

    SFA~\cite{fang2019selective}~
    & 0.700   & 0.607   & 0.793 & 0.647   & 0.840 	
    & 0.723   & 0.611   & 0.782 & 0.670   & 0.834 \\   

    PraNet~\cite{fan2020pra}~
    & 0.899   & 0.849   & 0.936 & 0.896   & 0.963  	
    & 0.898   & 0.840   & 0.915 & 0.885   & 0.944  \\  

    ACSNet~\cite{zhang2020adaptive}~
    & 0.882   & 0.826   & 0.927 & 0.873   & 0.947 	
    & 0.898   & 0.838   & 0.920 & 0.882   & 0.941  \\
    
    MSEG\cite{huang2021hardnet}~
    & 0.909   & 0.864   & 0.938 & 0.907   & 0.961 	
    & 0.897   & 0.839   & 0.912 & 0.885   & 0.942 \\

    DCRNet~\cite{yin2022duplex}~
    & 0.896   & 0.844   & 0.933 & 0.890   & 0.964 	
    & 0.886   & 0.825   & 0.911 & 0.868   & 0.933 \\   

    EU-Net~\cite{patel2021enhanced}~
    & 0.902   & 0.846   & 0.936 & 0.891   & 0.959  	
    & {0.908}   & {0.854}   & 0.917 & {0.893}   & {0.951}  \\  


    MSNet~\cite{zhao2021automatic}~
    & {0.918}   & {0.869}   & {0.946}  & {0.913}   & \textbf{0.973}  	
    & 0.905   & 0.849   & {0.923}  & 0.892   & 0.947 \\  
     

    CCBANet~\cite{nguyen2021ccbanet}~
    & 0.868   & 0.812   & 0.916   & 0.861   & 0.935  	
    & 0.853   & 0.777   & 0.887   & 0.831   & 0.901  \\  

    LDNet~\cite{zhang2022lesion}~
    & 0.881   & 0.825  & 0.924 & 0.879   & 0.960  	
    & 0.887   & 0.821  & 0.905 & 0.869   & 0.941  \\  
    
    \hline

    \textbf{EFA-Net}~
    & \textbf{0.919}   & \textbf{0.871}  & \textbf{0.943} & \textbf{0.916}   & {0.972}  	
    & \textbf{0.914}   & \textbf{0.861}  & \textbf{0.929} & \textbf{0.906}   & \textbf{0.955} \\

   \hline
  \end{tabular}\label{tab1}
\end{table*}

\begin{table*}[!t]
  \centering
  \renewcommand{\arraystretch}{1.1}
  \setlength\tabcolsep{6.2pt}
  \caption{{Results comparison on \zhh{the two unseen datasets (\ie, CVC-300 and ColonDB)}}.
  } 
  \small
  \begin{tabular}{r||ccccc|ccccc}
   \hline
    \multirow{2}*{\textbf{Methods}}   
    &\multicolumn{5}{c|}{CVC-300 \cite{vazquez2017benchmark}}
    &\multicolumn{5}{c}{ColonDB~\cite{tajbakhsh2015automated}}\\

     \cline{2-11}
  
    & mDice   & mIou   & $S_{\alpha}$  &$F_{\beta}^{w}$ & $E_{\phi}^{mean}$ 
    & mDice   & mIou   & $S_{\alpha}$  &$F_{\beta}^{w}$ & $E_{\phi}^{mean}$   \\

    \hline
    UNet~\cite{ronneberger2015u}~
    & 0.710   & 0.627   & 0.843 & 0.684   & 0.848 	
    & 0.504   & 0.436   & 0.710 & 0.491   & 0.692  \\
    
    UNet++~\cite{zhou2019unet++}~
    & 0.707   & 0.624   & 0.839 & 0.687   & 0.834 	
    & 0.482   & 0.408   & 0.693 & 0.467   & 0.680 \\

    SFA~\cite{fang2019selective}~
    & 0.467   & 0.329   & 0.640 & 0.341   & 0.644 	
    & 0.456   & 0.337   & 0.629 & 0.366   & 0.661 \\   

    PraNet~\cite{fan2020pra}~
    & 0.871   & 0.797   & 0.925 & 0.843   & 0.950  	
    & 0.712   & 0.640   & 0.820 & 0.699   & 0.847  \\  

    ACSNet~\cite{zhang2020adaptive}~
    & 0.863   & 0.787   & 0.923 & 0.825   & 0.939 	
    & 0.716   & 0.649   & 0.829 & 0.697   & 0.839  \\
    
    MSEG~\cite{huang2021hardnet}~
    & 0.874   & 0.804   & 0.924 & 0.852   & 0.948 	
    & 0.735   & 0.666   & 0.834 & 0.724   & 0.859 \\

    DCRNet~\cite{yin2022duplex}~
    & 0.856   & 0.788   & 0.921 & 0.830   & 0.943 	
    & 0.704   & 0.631   & 0.821 & 0.684   & 0.840 \\   

    EU-Net~\cite{patel2021enhanced}~
    & 0.837   & 0.765   & 0.904 & 0.805   & 0.919  	
    & {0.756}   & {0.681}   & 0.831 & 0.730   & 0.863 \\      
    

    MSNet~\cite{zhao2021automatic}~
    & 0.865   & 0.799   & 0.926 & 0.848   & 0.945  	
    & 0.751   & 0.671   & {0.838} & {0.736}   & 0.872  \\ 
    
        
    CCBANet~\cite{nguyen2021ccbanet}~
    & {0.888}   & {0.815}  & {0.935} & {0.862}   & \textbf{0.964}  
    & 0.706   & 0.626  & 0.812 & 0.676   & 0.852 \\  

    LDNet~\cite{zhang2022lesion}~
    & 0.869   & 0.793  & 0.923 & 0.841   & 0.948  
    & 0.740   & 0.652  & 0.830 & 0.717   & {0.876}  \\  

    \hline

    \textbf{EFA-Net}~
    & \textbf{0.894}   & \textbf{0.830}  & \textbf{0.941} & \textbf{0.878}   & {0.961}  
    & \textbf{0.774}   & \textbf{0.696}  & \textbf{0.855} & \textbf{0.753}   & \textbf{0.884} \\  
    
   \hline
  \end{tabular}\label{tab2}
\end{table*}

\section{Experiments and Results}
\label{sec4}

In this section, we first provide the details of the experimental settings, including datasets, evaluation metrics, and implementation details.
Then we report the segmentation results of our model and other comparison methods. Further, we carry out an ablation study to investigate the importance of each key component in the proposed segmentation model. 

\subsection{Experimental Settings}
\label{datasets}



\subsubsection{Datasets} We carry out the polyp segmentation task on five benchmark datasets, and the details of each dataset are introduced below. 

\begin{itemize}
\item \textbf{Kvasir}~\cite{jha2020kvasir}: This dataset is collected by Vestre Viken Health Trust in Norway from inside the gastrointestinal tract, which consists of $1,000$ colonoscopy images. 

\item \textbf{CVC-ClinicDB} \cite{bernal2015wm}: This dataset contains $612$ images collected from $29$ colonoscopy video sequences with a resolution of $288{\times}384$.

\item \textbf{CVC-ColonDB}~\cite{tajbakhsh2015automated}: This dataset consists of $380$ images with a resolution of $500\times{570}$. 

\item \textbf{ETIS}~\cite{silva2014toward}: This dataset consists of $196$ colonoscopy images with a size of $966\times{1225}$. 

\item \textbf{CVC-300}~\cite{vazquez2017benchmark}: This dataset includes $60$ colonoscopy images with a resolution of $500{\times}574$. 

\end{itemize}

Following the data partition in~\cite{fan2020pra}, $1,450$ images are randomly selected to construct the training set, where $900$ images are from Kvasir and $550$ ones are from CVC-ClinicDB. The remaining images from the two datasets (\ie, Kvasir and CVC-ClinicDB) and all other three datasets (\ie, CVC-ColonDB, ETIS, and CVC-300) are used for testing.

\subsubsection{Evaluation Metrics} To evaluate the effectiveness of our model, we first utilize two commonly adopted metrics \cite{yang2021mutual} in the segmentation task, namely mean dice score (Dice) and mean intersection over union (IoU).
Additionally, we adopt four popularly used metrics in the field of object detection \cite{zhou2021rgb,fan2020camouflaged}, \ie, S-measure ($S_{\alpha}$) \cite{fan2017structure}, F-measure \cite{achanta2009frequency} ($F_{\beta}^w$), $E_{\phi}^{mean}$ \cite{Fan2018Enhanced}, and precision-recall curve. 

\subsubsection{Implementation Details}
The proposed framework is implemented with PyTorch and trained on one NVIDIA Tesla P40 GPU with 24 GB memory. 
The Adam algorithm is adopted to optimize our framework with a learning rate of $1e$-$4$. 
To increase the diversity of input images, different data augmentation strategies are adopted, \ie, random flipping, crop, and rotation. Besides, we train the proposed model using three scaling ratios, \ie, $\{0.75, 1, 1.25\}$, and the resolutions of all inputs are uniformly rescaled to $352\times{352}$. 
Moreover, our framework is trained for $200$ epochs and the batch size is set to $16$. In the model inference stage, a test image is first resized to $352\times{352}$ and then fed into the network to obtain the predicted map, which is rescaled to its original size to conduct the final evaluation.


\begin{figure*}[!t]
\centering
\begin{overpic}[width=1.0\linewidth]{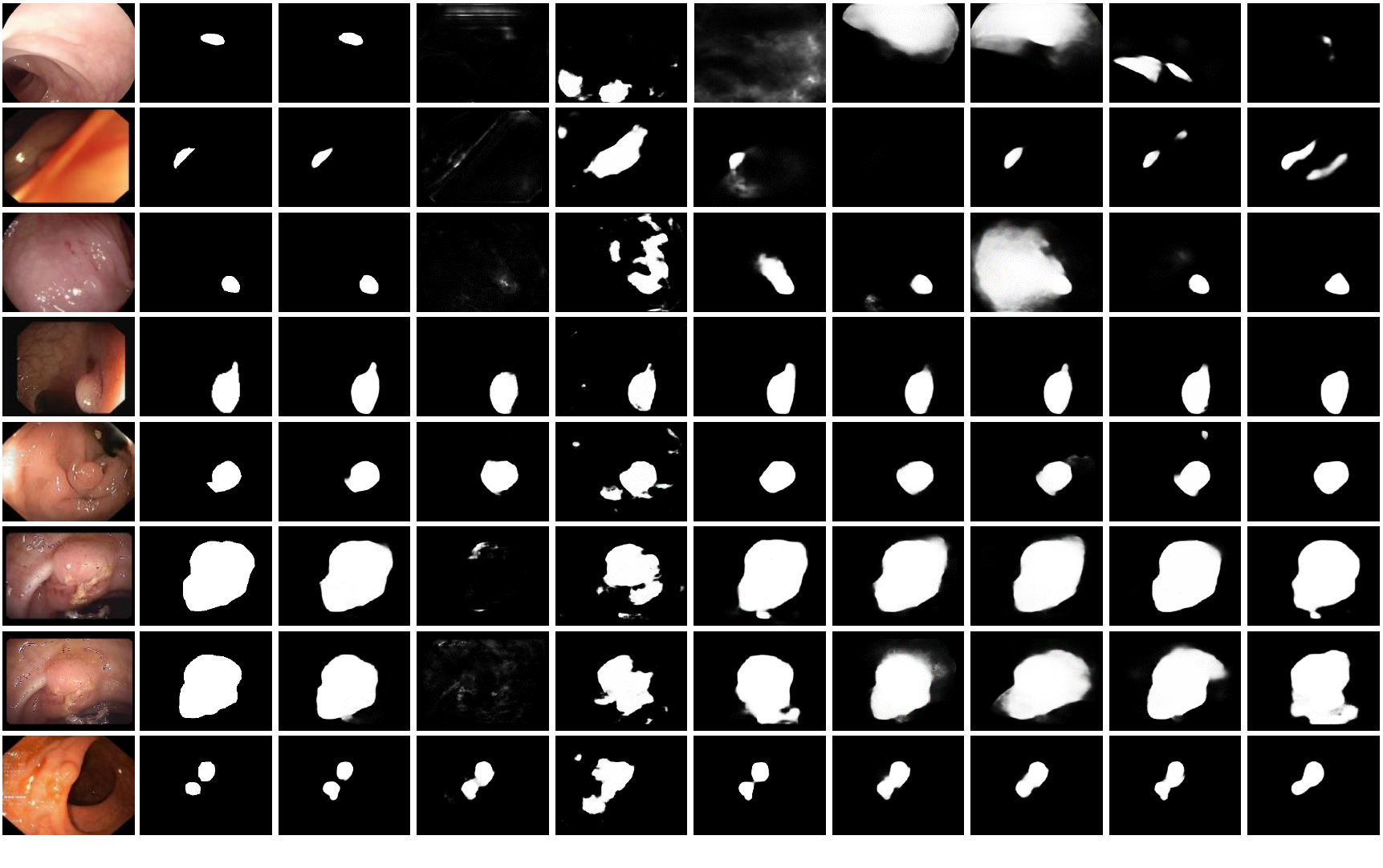}
\put(3, -0.15){\scriptsize Image}  
\put(14,-0.15){\scriptsize GT}  
\put(23,-0.15){\scriptsize Ours}  
\put(31,-0.15){\scriptsize UNet++}  
\put(43,-0.15){\scriptsize SFA}  
\put(52,-0.15){\scriptsize PraNet}  
\put(61,-0.15){\scriptsize ACSNet}  
\put(71,-0.15){\scriptsize DCRNet}  
\put(83,-0.15){\scriptsize MSNet}  
\put(92,-0.15){\scriptsize LDNet}   
\end{overpic}
\caption{{Qualitative results of our model and seven representative methods}.} 
 \label{fig5}
\end{figure*}




\begin{table}[!t]
  \centering
  \renewcommand{\arraystretch}{1.1}
  \setlength\tabcolsep{2.8pt}
  \caption{\zhh{Results comparison on the unseen ETIS dataset}.
  } 
  \small
  \begin{tabular}{r||ccccc}
   \hline

  
    Methods & mDice   & mIou   & $S_{\alpha}$  &$F_{\beta}^{w}$ & $E_{\phi}^{mean}$ \\

    \hline
    UNet~\cite{ronneberger2015u}~
    & 0.398   & 0.335   & 0.684  & 0.366  & 0.643	\\
    
    UNet++~\cite{zhou2019unet++}~
    & 0.401   & 0.344  & 0.683 & 0.390  & 0.629	 \\

    SFA~\cite{fang2019selective}~
    & 0.297  & 0.217  & 0.557 & 0.231  & 0.531	\\   

    PraNet~\cite{fan2020pra}~
    & 0.628  & 0.567  & 0.794 & 0.600  & 0.808 	 \\  

    ACSNet~\cite{zhang2020adaptive}~
    & 0.578   & 0.509   & 0.754 & 0.530   & 0.737 	 \\
    
    MSEG~\cite{huang2021hardnet}~
    & 0.700   & 0.630   & 0.828 & 0.671   & 0.855 	\\

    DCRNet~\cite{yin2022duplex}~
    & 0.556   & 0.496   & 0.736 & 0.506   & 0.742 	 \\   

    EU-Net~\cite{patel2021enhanced}~
    & 0.687   & 0.609   & 0.793 & 0.636   & 0.807  	\\      

    MSNet~\cite{zhao2021automatic}~
    & 0.723   & 0.652   & 0.845 & 0.677   & \textbf{0.875}  	 \\ 
    
    CCBANet~\cite{nguyen2021ccbanet}~
    & {0.559}   & {0.483}  & {0.751} & {0.513}   & 0.783 \\  

    LDNet~\cite{zhang2022lesion}~
    & 0.645   & 0.551  & 0.788 & 0.600   & 0.841   \\  

    \hline

    \textbf{EFA-Net}~
    & \textbf{0.749}   & \textbf{0.670}  & \textbf{0.858} & \textbf{0.698}   & {0.872}   \\  
    
   \hline
  \end{tabular}\label{tab3}
\end{table}

\begin{figure*}[!t]
	\begin{centering}
		\includegraphics[width=0.65\textwidth]{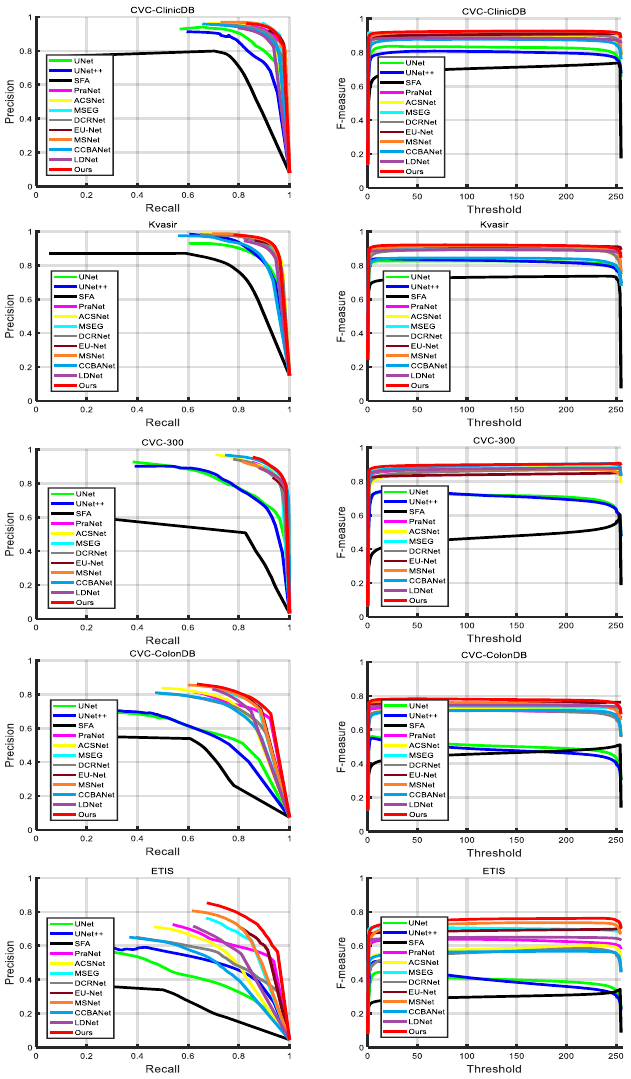}
		\caption{Precision-recall and F-measure curves on five colonoscopy datasets.}
		\label{fig6}
	\end{centering}
\end{figure*}

\begin{figure*}[!t]
	\begin{centering}
		\includegraphics[width=0.99\textwidth]{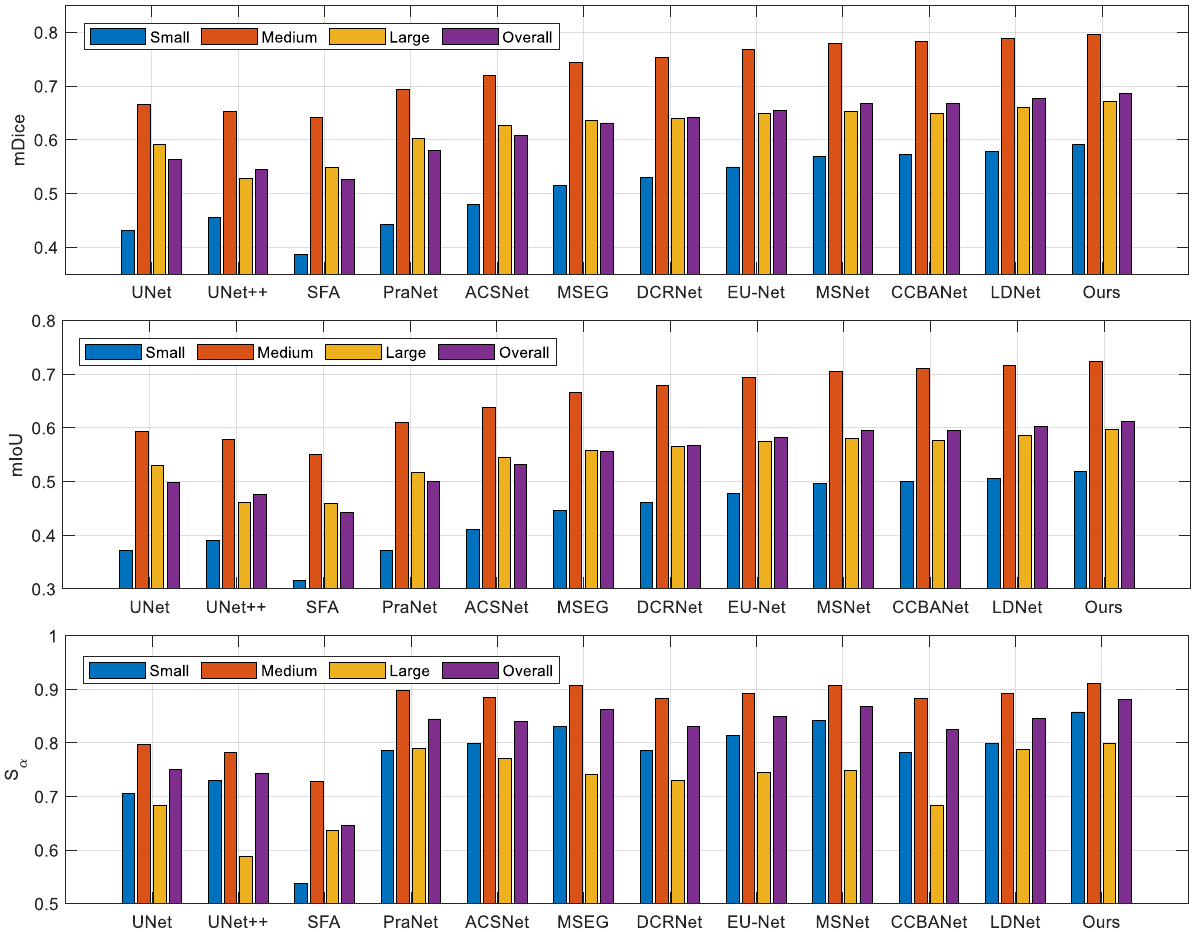}
		\caption{Scale-based evaluation \emph{w.r.t} different sizes of polyps.}
		\label{fig7}
	\end{centering}
\end{figure*}

\subsection{Segmentation Results}
\label{polyp}

\subsubsection{Comparison Baselines}
To validate the effectiveness of the proposed model, we compare EFA-Net with eleven state-of-the-art segmentation methods, \ie, UNet~\cite{ronneberger2015u}, UNet++~\cite{zhou2019unet++}, SFA~\cite{fang2019selective}, PraNet~\cite{fan2020pra}, ACSNet~\cite{zhang2020adaptive}, MSEG\cite{huang2021hardnet}, DCRNet~\cite{yin2022duplex}, EU-Net~\cite{patel2021enhanced}, 
MSNet~\cite{zhao2021automatic}, CCBANet~\cite{nguyen2021ccbanet}, and LDNet~\cite{zhang2022lesion}. For a fair comparison, we collect the segmentation maps from the original papers or run the released codes with default settings to obtain prediction maps.



\subsubsection{Quantitative Comparison}

Table~\ref{tab1} reports the quantitative results of different methods on the CVC-ClinicDB \cite{bernal2015wm} and Kvasir~\cite{jha2020kvasir} datasets. 
As shown in \tabref{tab1}, our EFA-Net outperforms all comparison approaches in most of the metrics. For example, EFA-Net achieves $4.3\%$, $4.2\%$, and $2.1\%$ improvements over LDNet in the terms of mDice, $F_{\beta}^{w}$, and $S_{\alpha}$ on the CVC-ClinicDB dataset. 
On the Kvasir dataset, our model achieves $3.0\%$, $4.9\%$, $2.7\%$, and $4.3\%$ improvements against LDNet in terms of mDice, mIou, $S_{\alpha}$, and $F_{\beta}^{w}$, respectively. In addition, our method achieves $7.2\%$, $10.8\%$, $4.7\%$, and $9.0\%$ improvements against CCBANet in terms of mDice, mIou, $S_{\alpha}$, and $F_{\beta}^{w}$. The results suggest that our EFA-Net accurately segments polyps under different challenges.
\zhh{This is mainly because our model can effectively capture multi-scale information and integrate cross-level and multi-scale features to capture the contextual semantic information, which yields better segmentation performance}. For three unseen datasets, our EFA-Net still performs better and achieves accurate segmentation than other compared methods as shown in \tabref{tab2} and \tabref{tab3}. \zhh{This further validates our EFA-Net has a better generalization ability than SOTA segmentation methods}. 

\zhh{It can be also noted that our method slightly lags behind certain approaches in terms of $E^{mean}_{\phi}$ on the CVC-ClinicDB, CVC-300, and ETIS datasets, but the differences are relatively small. This can be attributed to the fact that our method may overlook some finer details during the segmentation of large polyp regions, which are present in the aforementioned datasets. Therefore, we plan to focus on enhancing our model's ability to capture global and local context information. By incorporating these improvements, we aim to enhance the segmentation performance specifically for larger polyp regions in future research endeavors}.

Furthermore, we present precision-recall and F-measure curves in Fig.~\ref{fig6}. As shown Fig.~\ref{fig6}, it can be seen that our segmentation model achieves much better results compared to other state-of-the-art polyp segmentation methods. Thus, the results further validate the effectiveness of the proposed polyp segmentation method.

\subsubsection{Qualitative Comparison} 

Fig.~\ref{fig5} presents qualitative segmentation results of different methods. It can be observed that the results obtained by our EFA-Net are more similar to the ground truth, and our model outperforms other comparison methods in dealing with different challenging factors. 
For example, in the first two rows, the polyps have very small sizes, and our model still can accurately locate and segment them while some methods (\eg, UNet++ and SFA) completely fail to locate the polyp regions. In the $3^{rd}$ row, the polyp is visually embedded in its surrounding mucosa, while the boundary between the polyps and background is very blurred. In this case, our method can segment polyps more accurately than other comparison methods. It can be also noted that SFA, PraNet, and DCRNet produce several regions with over-segmented fragments. In the $4^{th}$ and $5^{th}$ rows, we observe that all methods can locate the main regions of the polyps, but our model obtains more accurate segmentation results. In the $6^{th}$ and $7^{th}$ rows, the polyps have relatively large sizes, making it challenging to accurately segment the complete regions of polyps. Under this challenge, our model still performs better than other methods, while PraNet and LDNet produce some non-polyp regions. \zhh{This is mainly because our scale-aware convolution module can learn scale-enhanced features by utilizing different dilated convolutions, in which different scale features are further aggregated to deal with scale variations. Consequently, our model effectively locates and segments polyps in different scales and also captures complete polyp regions. Moreover,  the incorporation of an edge-aware guidance module delivers crucial boundary cues to distinguish the semantic edges between the polyp and the background, which significantly improves segmentation performance.}
Furthermore, our method effectively segments multiple polyps (as the $8^{th}$ row). Overall, the qualitative comparison results further validate that our method achieves good performance in dealing with challenging factors for polyp segmentation. 

\subsubsection{Scale-based Evaluation} 

Scale variation is one of the most challenging factors in the polyp segmentation task. In this study, we have proposed to fully learn the scale-aware features and effectively fuse the cross-level features to enhance the discriminative ability of the learned features, which are helpful in dealing with scale variation. Thus, we conduct a scale-based evaluation for different methods. In this study, we calculate a ratio ($r$) involving the size of the polyp area into a given image to characterize the scale of the polyp region. To achieve this, three types of polyp scale are defined,\ie, 1) When $r<0.025$, the polyp is regarded as ``small" type; 2) When $r>0.2$, it is regarded as ``large” type; and 3) When the ratio is in the range of $[0.025, 0.2]$, we call it as ``medium” type. Further, we collect a hybrid dataset derived from the five used colonoscopy datasets (testing images), where $53.1\%$, $36.1\%$, and $10.8\%$ of images are with the three types, respectively. 
Fig.~\ref{fig7} presents the comparison results of the attribute-based study w.r.t. polyp scale in terms of three metrics (\ie, mDice, mIoU, and  $S_{\alpha}$). Based on the results, we have the following observations: 1) our proposed method obtains better performance over other compared methods under these conditions of different polyp scales, and 2) our model and all comparison methods achieve better performance in segmenting ``medium" polyps while they obtain relatively lower performance in segmenting other types. This suggests that scale variation is still a challenge for polyp segmentation, and it is relatively easy to segment the ``medium" polyps.

  \begin{table*}[t!]
  \centering
  \renewcommand{\arraystretch}{1.3}
  \renewcommand{\tabcolsep}{1.2mm}
  \caption{\zhh{Comparison of model parameters and FLOPs}.} 
  \scriptsize
  \begin{tabular}{r||c|c|c|c|c|c|c|c|c|c}
   \hline
    \textbf{Models}
    & UNet  & UNet++  & PraNet  & ACSNet & DCRNet & EU-Net & MSNet  & CCBANet  &LDNet  & EFA-Net \\
    
    \hline
    FLOPs         & 123.87    & 262.16  & 13.15  & 21.75 & 17.27  & 23.15   & 13.16   & 26.40 & 113.96 &33.20\\
    Param (M)     & 34.52     & 36.63   & 30.50  & 29.45 & 28.73  & 31.36   & 26.36   & 31.58 & 40.32  & 27.40 \\
    
  \hline

  \end{tabular}\label{tab42}
\end{table*}

\begin{table*}[tp]
  \centering
  \scriptsize
  \renewcommand{\arraystretch}{1.3}
  \setlength\tabcolsep{6.6pt}
  \caption{{Results of ablation study on the CVC-ClinicDB and Kvasir datasets}.} 
  \begin{tabular}{r||ccccc|ccccc}
  \hline
    
    \multirow{2}*{\textbf{Models~~}}   
    &\multicolumn{5}{c|}{CVC-ClinicDB \cite{bernal2015wm}}
    &\multicolumn{5}{c}{Kvasir~\cite{jha2020kvasir}}\\
    

    & mDice  & mIou  & $S_{\alpha}$  &$F_{\beta}^{w}$ & $E_{\phi}^{mean}$ 
    & mDice  & mIou  & $S_{\alpha}$  &$F_{\beta}^{w}$ & $E_{\phi}^{mean}$ \\

    \hline

    \emph{w/o} SCM ~
    & 0.910   & 0.865  & 0.939 & 0.907   & 0.966  	
    & 0.891   & 0.832  & 0.906 & 0.871   & 0.936 \\

    \emph{w/o} CFM~
    & 0.901   & 0.850  & 0.930 & 0.896   & 0.970  	
    & 0.897   & 0.840  & 0.914 & 0.880   & 0.944 \\  

    \emph{w/o} EGM~
    & 0.892   & 0.823  & 0.937 & 0.869   & 0.962  	
    & 0.898   & 0.842  & 0.917 & 0.884   & 0.940 \\  

    \hline

    \textbf{EFA-Net} ~
    & \textbf{0.919}   & \textbf{0.871}  & \textbf{0.943} & \textbf{0.916}   & \textbf{0.972}  	
    & \textbf{0.914}   & \textbf{0.861}  & \textbf{0.929} & \textbf{0.906}   & \textbf{0.955} \\  

   \hline
  \end{tabular}\label{tab5}
\end{table*}

\zhh{\subsubsection{Model Complexity Comparison}}

\zhh{To investigate the complexity of our model, we provide the model parameters for our method and other compared approaches in Table~\ref{tab42}. In Table~\ref{tab42}, $\#$Param is measured in million (M), and floating point operations (FLOPs) are measured in Giga (G). It can be observed that our model has relatively fewer parameters compared to other methods, indicating its efficiency in locating and segmenting polyps. However, it should be noted that our model requires more FLOPs than some of the comparison methods. This is primarily due to the utilization of the scale-aware convolution module for learning multi-scale feature representations and the cross-level fusion module for capturing local and global context information. Both of these modules involve various convolutional operations. Therefore, we will develop lightweight networks to improve the efficiency of the proposed model and achieve real-time polyp segmentation in future work.}

\subsection{Ablation Study}
\label{ablation}


This section investigates the importance of each module in the proposed EFA-Net, and the ablative results are shown in \tabref{tab5}. \zhh{In addition, we investigate the effects of different data augmentation strategies}.

\textbf{Effectiveness of SCM:} 
To validate the importance of SCM, we adopt a $1\times{1}$ convolutional layer instead of it, denoted as ``\emph{w/o} SCM". As reported in \tabref{tab5}, we observe that the performance of ``\emph{w/o} SCM" drops sharply on the Kvasir dataset. The results validate the effectiveness of SCM, which can learn scale-correlated features to deal with scale variation. 

\textbf{Effectiveness of CFM:} To investigate the contribution of CFM, we directly concatenate cross-level features, namely ``\emph{w/o} CFM". We can see that the performance decreases without using the proposed CFM, indicating the contribution of CFM that can fuse cross-level features to exploit both local and global context information for boosting polyp segmentation. 

\textbf{Effectiveness of EGM:} As shown in \tabref{tab5}, the results indicate that our EFA-Net achieves promising segmentation performance when incorporating the edge-aware feature into the decoder by the EGM.

\zhh{\textbf{Effectiveness of Data Augmentation:} To investigate the effects of different data augmentation strategies, we have conducted a series of experiments comparing them with a "Baseline" method that does not involve any data augmentation. The comparison results of different data augmentation strategies are presented in Table~\ref{tab06}. From the results, it is evident that our model achieves the highest performance when all three data augmentation strategies are employed. Conversely, the model's performance is notably poorer when no data augmentation is applied.}

\begin{table}[!t]
  \centering
  \renewcommand{\arraystretch}{1.2}
  \setlength\tabcolsep{2.2pt}
  \caption{\zhh{Effects of different data augmentation strategies on the Kvasir dataset}.
  } 
  \small
  \begin{tabular}{l||ccccc}
   \hline

  
    Strategies & mDice   & mIou   & $S_{\alpha}$  &$F_{\beta}^{w}$ & $E_{\phi}^{mean}$ \\

    \hline
    Baseline~~~~~~~
    & 0.899   & 0.841   & 0.914  & 0.889  & 0.946	\\
    
    + Crop
    & 0.905   & 0.848  & 0.919 & 0.894  & 0.950	 \\

    + Random flipping
    & 0.910  & 0.858  & 0.924 & 0.901  & 0.955	\\   

    + Rotation
    & 0.900  & 0.841  & 0.915 & 0.888  & 0.949 	 \\  

    All strategies
    & 0.914  & 0.861  & 0.929 & 0.906  & 0.955 	 \\  
    
   \hline
  \end{tabular}\label{tab06}
\end{table}

\zhh{\subsection{Failure Cases and Analyses}}

\begin{figure}[!t]
	\begin{centering}
		\includegraphics[width=0.48\textwidth]{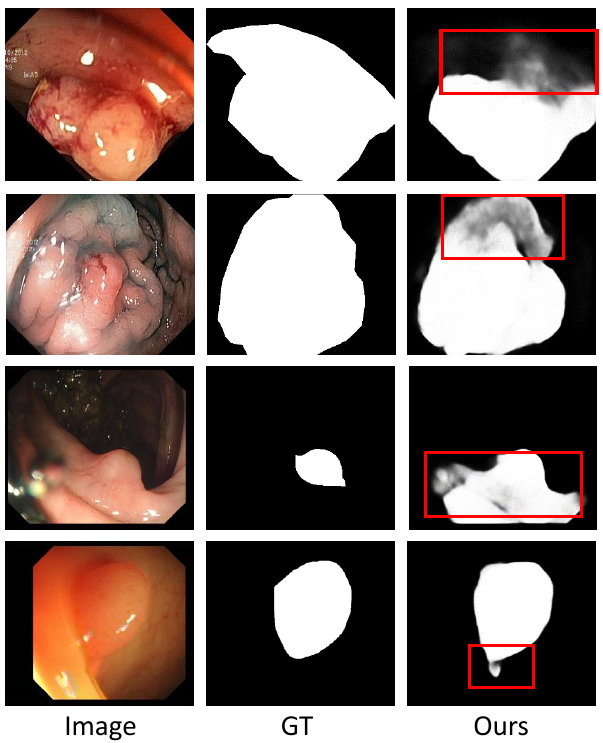}
		\caption{\zhh{Visualizations of some failure cases obtained by our model}.}
		\label{fig8}
	\end{centering}
\end{figure}

\zhh{The aforementioned experimental results demonstrate the superiority and effectiveness of the proposed EFA-Net. However, our model may struggle to accurately locate and segment polyp regions in complex scenarios such as those involving larger polyps or blurred boundaries. 
Visual representations of these failure cases are presented in Fig.~\ref{fig8}.
In the first two examples, the substantial size of the polyps complicates complete segmentation. As evident in these cases, our model fails to fully segment the entire polyp region, resulting in missing segments. 
In the $3^{rd}$ and $4^{th}$ rows, the boundaries between polyps and their surrounding mucosa are not sharp. In such instances, our EFA-Net overlooks the intricacies of boundary details or generates imprecise segmentation maps.
For future work, we aim to enhance EFA-Net to completely segment polyps of larger sizes, and harness boundary indicators more effectively, thereby improving overall segmentation performance. These developments will significantly contribute to the effectiveness of early detection and preventative measures for colorectal cancer}.

\zhh{\subsection{Discussion}}


\zhh{\textbf{Limitations}. As shown in Fig.~\ref{fig8}, our method fails to produce satisfactory segmentation maps when segmenting polyps on a large scale. To address this issue, we propose leveraging a large selective kernel network~\cite{li2023large} that dynamically adjusts its spatial receptive field to effectively capture the context of various objects under complex backgrounds. This approach helps to improve the segmentation performance for polyps in different sizes and variations.
Additionally, it's important to note that a real-time detection system with high accuracy is crucial in clinical practice. Such a system can assist doctors in taking necessary actions during colonoscopy procedures. However, our current model still requires significant computational resources, which may limit its practical application in clinical settings. To overcome this limitation, we plan to explore methods for model compression, such as network pruning \cite{molchanov2019importance} and knowledge distillation \cite{gou2021knowledge}. These techniques can be employed to develop a lightweight network specifically tailored for real-time polyp segmentation in clinics. By reducing the computational cost while maintaining sufficient accuracy, we aim to make our model more feasible and accessible for practical use in medical settings.} \vspace{0.3cm}

\zhh{As we are aware, noise and image artifacts are commonly encountered in medical images, making it crucial to address such scenarios that can significantly impact the clinical utility of the model. 
Firstly, it is beneficial to apply denoising algorithms or filters to the input images prior to feeding them into the segmentation model. This approach helps mitigate the effects of noise, enhance image quality, and reduce the impact of noise on the segmentation process.
Secondly, augmenting the training data with artificially generated noisy or artifact-laden images allows the model to learn and become more robust to variations present in real-world scenarios. This augmentation technique aids the model in generalizing better to different types of noise and artifacts during inference.
Thirdly, after obtaining the initial segmentation results, the application of post-processing techniques can refine the output. Such techniques may involve smoothing the boundaries, removing small artifacts, or enforcing geometric constraints to improve the overall segmentation performance.
Overall, our polyp segmentation model can effectively handle noise and image artifacts by employing a combination of denoising techniques, data augmentation, and post-processing strategies, thereby enhancing segmentation performance and clinical applicability.}

\zhh{Moreover, dealing with non-uniform textures and color variations in polyp segmentation poses a significant challenge due to the presence of different tissue types. To address these challenges, there are several solutions that can be discussed. 
Firstly, non-uniform textures and color variations can manifest at various scales within polyps. Therefore, incorporating multi-scale analysis allows the model to capture both local details and global context. This can be achieved by employing network architectures that include multiple branches or receptive fields of different sizes, enabling the model to adaptively capture information from various levels of granularity. 
Secondly, augmenting the training data with various transformations, such as color transformation, can enhance the model's ability to handle non-uniform textures and color variations. 
Thirdly, attention mechanisms can enhance the model's focus on informative regions while suppressing irrelevant or noisy information. By incorporating attention mechanisms into the network architecture, the model can attend to specific regions exhibiting non-uniform textures or color variations, improving its ability to segment polyps accurately.
Lastly, pre-training the model on large-scale datasets containing diverse images, including those with non-uniform textures and color variations, can provide a strong initial feature representation. Fine-tuning the model on the downstream polyp segmentation task enables it to generalize effectively to variations in textures and colors encountered during testing. Overall, by considering these approaches, the model's performance in handling non-uniform textures and color variations can be improved, leading to more accurate and robust polyp segmentation results}. \vspace{0.5cm}

\section{Conclusion}
\label{sec5}


In this paper, we have presented a novel EFA-Net framework for polyp segmentation in colonoscopy images. To explore edge semantic information, an edge-aware guidance module is proposed to learn the edge-enhanced feature and then fed it into the decoder network to benefit the segmentation performance. Moreover, the scale-aware convolution module and cross-level fusion module are proposed to capture multi-scale information and integrate cross-level features to handle the issue of scale variation. Extensive experimental results on benchmark datasets validate the effectiveness of our EFA-Net in accurately segmenting polyps.
More importantly, to promote future research for the polyp segmentation community, we will collect awesome polyp segmentation models at: \href{https://github.com/taozh2017/Awesome-Polyp-Segmentation}{https://github.com/taozh2017/Awesome-Polyp-Segmentation}.

%


%

%
%

\bibliographystyle{IEEEtran}
\bibliography{mir-article}

\begin{thebibliography}{10}
\providecommand{\url}[1]{#1}
\csname url@samestyle\endcsname
\providecommand{\newblock}{\relax}
\providecommand{\bibinfo}[2]{#2}
\providecommand{\BIBentrySTDinterwordspacing}{\spaceskip=0pt\relax}
\providecommand{\BIBentryALTinterwordstretchfactor}{4}
\providecommand{\BIBentryALTinterwordspacing}{\spaceskip=\fontdimen2\font plus
\BIBentryALTinterwordstretchfactor\fontdimen3\font minus \fontdimen4\font\relax}
\providecommand{\BIBforeignlanguage}[2]{{%
\expandafter\ifx\csname l@#1\endcsname\relax
\typeout{** WARNING: IEEEtran.bst: No hyphenation pattern has been}%
\typeout{** loaded for the language `#1'. Using the pattern for}%
\typeout{** the default language instead.}%
\else
\language=\csname l@#1\endcsname
\fi
#2}}
\providecommand{\BIBdecl}{\relax}
\BIBdecl

\bibitem{silva2014toward}
J.~Silva, A.~Histace, O.~Romain, X.~Dray, and B.~Granado, ``Toward embedded detection of polyps in wce images for early diagnosis of colorectal cancer,'' \emph{International Journal of Computer Assisted Radiology and Surgery}, vol.~9, no.~2, pp. 283--293, 2014, {DOI}: \href{10.1007/s11548-013-0926-3}{10.1007/s11548-013-0926-3}.

\bibitem{ronneberger2015u}
O.~Ronneberger, P.~Fischer, and T.~Brox, ``{U-Net: Convolutional networks for biomedical image segmentation},'' in \emph{{International Conference on Medical Image Computing and Computer Assisted Intervention}}.\hskip 1em plus 0.5em minus 0.4em\relax {Munich, Germany}: Springer, 2015, pp. 234--241, {{DOI}: \href{10.1007/978-3-319-24574-4_28}{10.1007/978-3-319-24574-4_28}}.

\bibitem{zhou2019unet++}
Z.~Zhou, M.~M.~R. Siddiquee, N.~Tajbakhsh, and J.~Liang, ``Unet++: Redesigning skip connections to exploit multiscale features in image segmentation,'' \emph{IEEE Transactions on Medical Imaging}, vol.~39, no.~6, pp. 1856--1867, 2019.

\bibitem{jha2019resunet++}
D.~Jha, P.~H. Smedsrud, M.~A. Riegler, D.~Johansen, T.~De~Lange, P.~Halvorsen, and H.~D. Johansen, ``Resunet++: An advanced architecture for medical image segmentation,'' in \emph{IEEE International Symposium on Multimedia}, 2019, pp. 225--2255.

\bibitem{patel2021enhanced}
K.~Patel, A.~M. Bur, and G.~Wang, ``Enhanced u-net: A feature enhancement network for polyp segmentation,'' in \emph{Proc. IEEE Int. Robots and Vision}, 2021, pp. 181--188.

\bibitem{zhang2020adaptive}
R.~Zhang, G.~Li, Z.~Li, S.~Cui, D.~Qian, and Y.~Yu, ``Adaptive context selection for polyp segmentation,'' in \emph{{International Conference on Medical Image Computing and Computer Assisted Intervention}}.\hskip 1em plus 0.5em minus 0.4em\relax {Lima, Peru}: Springer, 2020, pp. 253--262, {DOI}: \href{10.1007/978-3-030-59725-2_25}{10.1007/978-3-030-59725-2_25}.

\bibitem{nguyen2021ccbanet}
T.-C. Nguyen, T.-P. Nguyen, G.-H. Diep, A.-H. Tran-Dinh, T.~V. Nguyen, and M.-T. Tran, ``Ccbanet: Cascading context and balancing attention for polyp segmentation,'' in \emph{{International Conference on Medical Image Computing and Computer Assisted Intervention}}.\hskip 1em plus 0.5em minus 0.4em\relax {Strasbourg, France}: Springer, 2021, pp. 633--643, {DOI}: \href{10.1007/978-3-030-87193-2_60}{10.1007/978-3-030-87193-2_60}.

\bibitem{murugesan2019psi}
B.~Murugesan, K.~Sarveswaran, S.~M. Shankaranarayana, K.~Ram, J.~Joseph, and M.~Sivaprakasam, ``Psi-{N}et: Shape and boundary aware joint multi-task deep network for medical image segmentation,'' in \emph{Engineering in Medicine and Biology Society}.\hskip 1em plus 0.5em minus 0.4em\relax Germany, Germany: IEEE, 2019, pp. 7223--7226, {DOI}: \href{10.1109/EMBC.2019.8857339}{10.1109/EMBC.2019.8857339}.

\bibitem{fang2019selective}
Y.~Fang, C.~Chen, Y.~Yuan, and K.-y. Tong, ``Selective feature aggregation network with area-boundary constraints for polyp segmentation,'' in \emph{{International Conference on Medical Image Computing and Computer Assisted Intervention}}.\hskip 1em plus 0.5em minus 0.4em\relax {Shenzhen, China}: Springer, 2019, pp. 302--310, {DOI}: \href{10.1007/978-3-030-32239-7_34}{10.1007/978-3-030-32239-7_34}.

\bibitem{fan2020pra}
D.-P. Fan, G.-P. Ji, T.~Zhou, G.~Chen, H.~Fu, J.~Shen, and L.~Shao, ``Pranet: Parallel reverse attention network for polyp segmentation,'' in \emph{{International Conference on Medical Image Computing and Computer Assisted Intervention}}.\hskip 1em plus 0.5em minus 0.4em\relax {Lima, Peru}: Springer, 2020, pp. 263--273.

\bibitem{hwang2007polyp}
S.~Hwang, J.~Oh, W.~Tavanapong, J.~Wong, and P.~C. De~Groen, ``Polyp detection in colonoscopy video using elliptical shape feature,'' in \emph{IEEE International Conference on Image Processing}, vol.~2.\hskip 1em plus 0.5em minus 0.4em\relax IEEE, 2007, pp. II--465.

\bibitem{bernal2012towards}
J.~Bernal, J.~S{\'a}nchez, and F.~Vilarino, ``Towards automatic polyp detection with a polyp appearance model,'' \emph{{Pattern Recognition}}, vol.~45, no.~9, pp. 3166--3182, 2012, {DOI}: \href{10.1016/j.patcog.2012.03.002}{10.1016/j.patcog.2012.03.002}.

\bibitem{wang2013part}
Y.~Wang, W.~Tavanapong, J.~Wong, J.~Oh, and P.~C. De~Groen, ``Part-based multiderivative edge cross-sectional profiles for polyp detection in colonoscopy,'' \emph{IEEE Journal of Biomedical and Health Informatics}, vol.~18, no.~4, pp. 1379--1389, 2013.

\bibitem{mamonov2014automated}
A.~V. Mamonov, I.~N. Figueiredo, P.~N. Figueiredo, and Y.-H.~R. Tsai, ``Automated polyp detection in colon capsule endoscopy,'' \emph{IEEE Transactions on Medical Imaging}, vol.~33, no.~7, pp. 1488--1502, 2014, {DOI}: \href{10.1109/TMI.2014.2314959}{10.1109/TMI.2014.2314959}.

\bibitem{tajbakhsh2015automated}
N.~{Tajbakhsh}, S.~R. {Gurudu}, and J.~{Liang}, ``Automated polyp detection in colonoscopy videos using shape and context information,'' \emph{IEEE Transactions on Medical Imaging}, vol.~35, no.~2, pp. 630--644, 2016, {DOI}: \href{10.1109/TMI.2015.2487997}{10.1109/TMI.2015.2487997}.

\bibitem{li2017colorectal}
Q.~Li, G.~Yang, Z.~Chen, B.~Huang, L.~Chen, D.~Xu, and OTHERS, ``Colorectal polyp segmentation using a fully convolutional neural network,'' in \emph{IEEE International Congress on Image and Signal Processing, Biomedical Engineering and Informatics}.\hskip 1em plus 0.5em minus 0.4em\relax IEEE, 2017, pp. 1--5.

\bibitem{ibtehaz2020multiresunet}
N.~Ibtehaz and M.~S. Rahman, ``Multiresunet: Rethinking the u-net architecture for multimodal biomedical image segmentation,'' \emph{Neural Networks}, vol. 121, pp. 74--87, 2020.

\bibitem{zhang2018road}
Z.~Zhang, Q.~Liu, and Y.~Wang, ``Road extraction by deep residual u-net,'' \emph{IEEE Geoscience and Remote Sensing Letters}, vol.~15, no.~5, pp. 749--753, 2018.

\bibitem{li2018h}
X.~Li, H.~Chen, X.~Qi, Q.~Dou, C.-W. Fu, and P.-A. Heng, ``H-{D}ense{UN}et: hybrid densely connected unet for liver and tumor segmentation from ct volumes,'' \emph{IEEE Transactions on Medical Imaging}, vol.~37, no.~12, pp. 2663--2674, 2018.

\bibitem{oktay2018attention}
O.~Oktay, J.~Schlemper, L.~L. Folgoc, M.~Lee, M.~Heinrich, K.~Misawa, K.~Mori, S.~McDonagh, N.~Y. Hammerla, B.~Kainz \emph{et~al.}, ``Attention u-net: Learning where to look for the pancreas,'' \emph{arXiv preprint arXiv:1804.03999}, 2018.

\bibitem{zhoupolyp2023}
T.~Zhou, Y.~Zhou, K.~He, C.~Gong, J.~Yang, H.~Fu, and D.~Shen, ``Cross-level feature aggregation network for polyp segmentation,'' \emph{Pattern Recognition}, vol. 140, p. 109555, 2023.

\bibitem{yue2022boundary}
G.~Yue, W.~Han, B.~Jiang, T.~Zhou, R.~Cong, and T.~Wang, ``Boundary constraint network with cross layer feature integration for polyp segmentation,'' \emph{IEEE Journal of Biomedical and Health Informatics}, vol.~26, no.~8, pp. 4090--4099, 2022.

\bibitem{liu2022dbmf}
F.~Liu, Z.~Hua, J.~Li, and L.~Fan, ``Dbmf: Dual branch multiscale feature fusion network for polyp segmentation,'' \emph{Computers in Biology and Medicine}, vol. 151, p. 106304, 2022.

\bibitem{su2023accurate}
Y.~Su, J.~Cheng, C.~Zhong, C.~Jiang, J.~Ye, and J.~He, ``Accurate polyp segmentation through enhancing feature fusion and boosting boundary performance,'' \emph{Neurocomputing}, vol. 545, p. 126233, 2023.

\bibitem{song2022attention}
P.~Song, J.~Li, and H.~Fan, ``Attention based multi-scale parallel network for polyp segmentation,'' \emph{Computers in Biology and Medicine}, vol. 146, p. 105476, 2022.

\bibitem{tomar2022tganet}
N.~K. Tomar, D.~Jha, U.~Bagci, and S.~Ali, ``Tganet: text-guided attention for improved polyp segmentation,'' in \emph{{International Conference on Medical Image Computing and Computer Assisted Intervention}}.\hskip 1em plus 0.5em minus 0.4em\relax Springer, 2022, pp. 151--160.

\bibitem{wei2021shallow}
J.~Wei, Y.~Hu, R.~Zhang, Z.~Li, S.~K. Zhou, and S.~Cui, ``Shallow attention network for polyp segmentation,'' in \emph{{International Conference on Medical Image Computing and Computer Assisted Intervention}}.\hskip 1em plus 0.5em minus 0.4em\relax {Strasbourg, France}: Springer, 2021, pp. 699--708, {DOI}: \href{10.1007/978-3-030-87193-2_66}{10.1007/978-3-030-87193-2_66}.

\bibitem{wu2022polypseg}
H.~Wu, Z.~Zhao, J.~Zhong, W.~Wang, Z.~Wen, and J.~Qin, ``Polypseg+: A lightweight context-aware network for real-time polyp segmentation,'' \emph{{IEEE Transactions on Cybernetics}}, 2022.

\bibitem{yin2022duplex}
Z.~Yin, K.~Liang, Z.~Ma, and J.~Guo, ``Duplex contextual relation network for polyp segmentation,'' in \emph{Proc. Int. Symp. Biomed. Imaging}, 2022, pp. 1--5.

\bibitem{lin2022bsca}
Y.~Lin, J.~Wu, G.~Xiao, J.~Guo, G.~Chen, and J.~Ma, ``Bsca-net: Bit slicing context attention network for polyp segmentation,'' \emph{Pattern Recognition}, vol. 132, p. 108917, 2022.

\bibitem{ding2019boundary}
H.~Ding, X.~Jiang, A.~Q. Liu, N.~M. Thalmann, and G.~Wang, ``Boundary-aware feature propagation for scene segmentation,'' in \emph{Proceedings of the IEEE International Conference on Computer Vision}, 2019, pp. 6819--6829.

\bibitem{qin2019basnet}
X.~Qin, Z.~Zhang, C.~Huang, C.~Gao, M.~Dehghan, and M.~Jagersand, ``Basnet: Boundary-aware salient object detection,'' in \emph{{Conference on Computer Vision and Pattern Recognition}}, 2019, pp. 7479--7489.

\bibitem{liu2022weakly}
K.~Liu, Y.~Zhao, Q.~Nie, Z.~Gao, and B.~M. Chen, ``Weakly supervised 3d scene segmentation with region-level boundary awareness and instance discrimination,'' in \emph{European Conference on Computer V ision}.\hskip 1em plus 0.5em minus 0.4em\relax Springer, 2022, pp. 37--55.

\bibitem{zhou2022feature}
T.~Zhou, Y.~Zhou, C.~Gong, J.~Yang, and Y.~Zhang, ``Feature aggregation and propagation network for camouflaged object detection,'' \emph{IEEE Trans. Image Process.}, vol.~31, pp. 7036--7047, 2022.

\bibitem{li2023image}
F.~Li, X.~Du, L.~Zhang, and A.~Liu, ``Image feature fusion method based on edge detection,'' \emph{Information Technology and Control}, vol.~52, no.~1, pp. 5--24, 2023.

\bibitem{nawaz2022melanoma}
M.~Nawaz, T.~Nazir, M.~Masood, F.~Ali, M.~A. Khan, U.~Tariq, N.~Sahar, and R.~Dama{\v{s}}evi{\v{c}}ius, ``Melanoma segmentation: A framework of improved densenet77 and unet convolutional neural network,'' \emph{International Journal of Imaging Systems and Technology}, vol.~32, no.~6, pp. 2137--2153, 2022.

\bibitem{maskeliunas2023pareto}
R.~Maskeliunas, R.~Damasevicius, D.~Vitkute-Adzgauskiene, and S.~Misra, ``Pareto optimized large mask approach for efficient and background humanoid shape removal,'' \emph{IEEE access}, 2023.

\bibitem{chen2017dcan}
H.~Chen, X.~Qi, L.~Yu, Q.~Dou, J.~Qin, and P.-A. Heng, ``Dcan: Deep contour-aware networks for object instance segmentation from histology images,'' \emph{Medical Image Analysis}, vol.~36, pp. 135--146, 2017.

\bibitem{zhang2019net}
Z.~Zhang, H.~Fu, H.~Dai, J.~Shen, Y.~Pang, and L.~Shao, ``Et-net: A generic edge-attention guidance network for medical image segmentation,'' in \emph{{International Conference on Medical Image Computing and Computer Assisted Intervention}}.\hskip 1em plus 0.5em minus 0.4em\relax Springer, 2019, pp. 442--450.

\bibitem{ramasamy2021detection}
L.~K. Ramasamy, S.~G. Padinjappurathu, S.~Kadry, and R.~Dama{\v{s}}evi{\v{c}}ius, ``Detection of diabetic retinopathy using a fusion of textural and ridgelet features of retinal images and sequential minimal optimization classifier,'' \emph{PeerJ computer science}, vol.~7, p. e456, 2021.

\bibitem{wang2022boundary}
R.~Wang, S.~Chen, C.~Ji, J.~Fan, and Y.~Li, ``Boundary-aware context neural network for medical image segmentation,'' \emph{Medical Image Analysis}, vol.~78, p. 102395, 2022, {DOI}: \href{10.1016/j.media.2022.102395}{10.1016/j.media.2022.102395}.

\bibitem{wang2022eanet}
K.~Wang, X.~Zhang, X.~Zhang, Y.~Lu, S.~Huang, and D.~Yang, ``{EAN}et: Iterative edge attention network for medical image segmentation,'' \emph{Pattern Recognition}, vol. 127, p. 108636, 2022.

\bibitem{kim2018san}
Y.~Kim, B.-N. Kang, and D.~Kim, ``San: Learning relationship between convolutional features for multi-scale object detection,'' in \emph{Proceedings of the European Conference on Computer Vision}, 2018, pp. 316--331.

\bibitem{cao2019high}
J.~Cao, Y.~Pang, S.~Zhao, and X.~Li, ``High-level semantic networks for multi-scale object detection,'' \emph{IEEE Transactions on Circuits and Systems for Video Technology}, vol.~30, no.~10, pp. 3372--3386, 2019.

\bibitem{zhoucvmj2023}
T.~Zhou, D.-P. Fan, G.~Chen, Y.~Zhou, and H.~Fu, ``Specificity-preserving rgb-d saliency detection,'' \emph{Computational Visual Media}, vol.~9, pp. 297--317, 2023.

\bibitem{li2018multi}
J.~Li, F.~Fang, K.~Mei, and G.~Zhang, ``Multi-scale residual network for image super-resolution,'' in \emph{Proceedings of the European Conference on Computer Vision}, 2018, pp. 517--532.

\bibitem{li2020mdcn}
J.~Li, F.~Fang, J.~Li, K.~Mei, and G.~Zhang, ``{MDCN}: {M}ulti-scale dense cross network for image super-resolution,'' \emph{IEEE Transactions on Circuits and Systems for Video Technology}, vol.~31, no.~7, pp. 2547--2561, 2020.

\bibitem{he2019dynamic}
J.~He, Z.~Deng, and Y.~Qiao, ``Dynamic multi-scale filters for semantic segmentation,'' in \emph{Proceedings of the IEEE/CVF International Conference on Computer Vision}, 2019, pp. 3562--3572.

\bibitem{gu2022multi}
J.~Gu, H.~Kwon, D.~Wang, W.~Ye, M.~Li, Y.-H. Chen, L.~Lai, V.~Chandra, and D.~Z. Pan, ``Multi-scale high-resolution vision transformer for semantic segmentation,'' in \emph{Proceedings of the IEEE/CVF Conference on Computer Vision and Pattern Recognition}, 2022, pp. 12\,094--12\,103.

\bibitem{zhao2017pyramid}
H.~Zhao, J.~Shi, X.~Qi, X.~Wang, and J.~Jia, ``Pyramid scene parsing network,'' in \emph{Proc. IEEE Conf. Comput. Vision Pattern Recognit.}, 2017, pp. 2881--2890.

\bibitem{chen2017deeplab}
L.-C. Chen, G.~Papandreou, I.~Kokkinos, K.~Murphy, and A.~L. Yuille, ``Deeplab: Semantic image segmentation with deep convolutional nets, atrous convolution, and fully connected crfs,'' \emph{IEEE Trans. Pattern Anal. Mach. Intell.}, vol.~40, no.~4, pp. 834--848, 2017.

\bibitem{yang2019clci}
H.~Yang, W.~Huang, K.~Qi, C.~Li, X.~Liu, M.~Wang, H.~Zheng, and S.~Wang, ``{CLCI-N}et: {C}ross-level fusion and context inference networks for lesion segmentation of chronic stroke,'' in \emph{Medical Image Computing and Computer Assisted Intervention}.\hskip 1em plus 0.5em minus 0.4em\relax Springer, 2019, pp. 266--274.

\bibitem{zhao2021automatic}
X.~Zhao, L.~Zhang, and H.~Lu, ``Automatic polyp segmentation via multi-scale subtraction network,'' in \emph{{International Conference on Medical Image Computing and Computer Assisted Intervention}}.\hskip 1em plus 0.5em minus 0.4em\relax {Strasbourg, France}: Springer, 2021, pp. 120--130, {{DOI}: \href{10.1007/978-3-030-87193-2_12}{10.1007/978-3-030-87193-2_12}}.

\bibitem{srivastava2021msrf}
A.~Srivastava, D.~Jha, S.~Chanda, U.~Pal, H.~D. Johansen \emph{et~al.}, ``Msrf-net: A multi-scale residual fusion network for biomedical image segmentation,'' \emph{IEEE Journal of Biomedical and Health Informatics}, vol.~26, no.~5, pp. 2252--2263, 2021.

\bibitem{yang2023flexible}
H.~Yang, T.~Zhou, Y.~Zhou, Y.~Zhang, and H.~Fu, ``Flexible fusion network for multi-modal brain tumor segmentation,'' \emph{IEEE Journal of Biomedical and Health Informatics}, 2023.

\bibitem{pami20Res2net}
S.-H. Gao, M.-M. Cheng, K.~Zhao, X.-Y. Zhang, M.-H. Yang, and P.~Torr, ``Res2net: A new multi-scale backbone architecture,'' \emph{{IEEE Transactions on Pattern Analysis and Machine Intelligence}}, vol.~43, no.~2, pp. 652--662, 2019, {DOI}: \href{10.1109/TPAMI.2019.2938758}{10.1109/TPAMI.2019.2938758}.

\bibitem{fan2020inf}
D.-P. Fan, T.~Zhou, G.-P. Ji, Y.~Zhou, G.~Chen, H.~Fu, J.~Shen, and L.~Shao, ``Inf-net: Automatic covid-19 lung infection segmentation from ct images,'' \emph{IEEE Transactions on Medical Imaging}, vol.~39, no.~8, pp. 2626--2637, 2020, {{DOI}: \href{10.1109/TMI.2020.2996645}{10.1109/TMI.2020.2996645}}.

\bibitem{sun2022boundary}
Y.~Sun, S.~Wang, C.~Chen, and T.-Z. Xiang, ``Boundary-guided camouflaged object detection,'' in \emph{{International Joint Conference on Artificial Intelligence}}, 2022, pp. 1335--1341.

\bibitem{zhao2019egnet}
J.-X. Zhao, J.-J. Liu, D.-P. Fan, Y.~Cao, J.~Yang, and M.-M. Cheng, ``{EGNet: Edge guidance network for salient object detection},'' in \emph{{International Conference on Computer Vision}}, 2019, pp. 8779--8788.

\bibitem{dai2021attentional}
Y.~Dai, F.~Gieseke, S.~Oehmcke, Y.~Wu, and K.~Barnard, ``Attentional feature fusion,'' in \emph{IEEE WACV}, 2021, pp. 3560--3569.

\bibitem{wei2019f3net}
J.~Wei, S.~Wang, and Q.~Huang, ``{F3Net: Fusion, Feedback and Focus for Salient Object Detection},'' in \emph{{AAAI Conference on Artificial Intelligence}}, 2020, pp. 12\,321--12\,328.

\bibitem{bernal2015wm}
J.~Bernal, F.~J. S{\'a}nchez, G.~Fern{\'a}ndez-Esparrach, D.~Gil, C.~Rodr{\'\i}guez, and F.~Vilari{\~n}o, ``Wm-dova maps for accurate polyp highlighting in colonoscopy: Validation vs. saliency maps from physicians,'' \emph{Computerized Medical Imaging and Graphics}, vol.~43, pp. 99--111, 2015, {DOI}: \href{10.1016/j.compmedimag.2015.02.007}{10.1016/j.compmedimag.2015.02.007}.

\bibitem{jha2020kvasir}
D.~Jha, P.~H. Smedsrud, M.~A. Riegler, P.~Halvorsen, T.~de~Lange, D.~Johansen, and H.~D. Johansen, ``Kvasir-seg: A segmented polyp dataset,'' in \emph{International Conference on Multimedia Modeling}.\hskip 1em plus 0.5em minus 0.4em\relax {Daejeon, Korea}: Springer, 2020, pp. 451--462.

\bibitem{huang2021hardnet}
C.-H. Huang, H.-Y. Wu, and Y.-L. Lin, ``Hardnet-mseg: A simple encoder-decoder polyp segmentation neural network that achieves over 0.9 mean dice and 86 fps,'' \emph{arXiv preprint arXiv:2101.07172}, 2021.

\bibitem{zhang2022lesion}
R.~Zhang, P.~Lai, X.~Wan, D.-J. Fan, F.~Gao, X.-J. Wu, and G.~Li, ``Lesion-aware dynamic kernel for polyp segmentation,'' in \emph{{International Conference on Medical Image Computing and Computer Assisted Intervention}}.\hskip 1em plus 0.5em minus 0.4em\relax Springer, 2022, pp. 99--109.

\bibitem{vazquez2017benchmark}
D.~V{\'a}zquez, J.~Bernal, F.~J. S{\'a}nchez, G.~Fern{\'a}ndez-Esparrach, A.~M. L{\'o}pez, A.~Romero, M.~Drozdzal, and A.~Courville, ``A benchmark for endoluminal scene segmentation of colonoscopy images,'' \emph{Journal of Healthcare Engineering}, vol. 2017, p. 4037190, 2017, {DOI}: \href{10.1155/2017/4037190}{10.1155/2017/4037190}.

\bibitem{yang2021mutual}
C.~Yang, X.~Guo, M.~Zhu, B.~Ibragimov, and Y.~Yuan, ``Mutual-prototype adaptation for cross-domain polyp segmentation,'' \emph{IEEE J. Biomed. Health Informat.}, vol.~25, no.~10, pp. 3886--3897, 2021.

\bibitem{zhou2021rgb}
T.~Zhou, D.-P. Fan, M.-M. Cheng, J.~Shen, and L.~Shao, ``{RGB-D} salient object detection: {A} survey,'' \emph{Computational Visual Media}, vol.~7, no.~1, pp. 37--69, 2021.

\bibitem{fan2020camouflaged}
D.-P. Fan, G.-P. Ji, G.~Sun, M.-M. Cheng, J.~Shen, and L.~Shao, ``Camouflaged object detection,'' in \emph{{Conference on Computer Vision and Pattern Recognition}}.\hskip 1em plus 0.5em minus 0.4em\relax {Seattle, WA, USA}: IEEE, 2020, pp. 2777--2787, {{DOI}: \href{10.1109/CVPR42600.2020.00285}{10.1109/CVPR42600.2020.00285}}.

\bibitem{fan2017structure}
D.-P. Fan, M.-M. Cheng, Y.~Liu, T.~Li, and A.~Borji, ``Structure-measure: A new way to evaluate foreground maps,'' in \emph{{International Conference on Computer Vision}}.\hskip 1em plus 0.5em minus 0.4em\relax Venice, Italy: IEEE, 2017, pp. 4548--4557, {{DOI}: \href{10.1109/ICCV.2017.487}{10.1109/ICCV.2017.487}}.

\bibitem{achanta2009frequency}
R.~Achanta, S.~Hemami, F.~Estrada, and S.~Susstrunk, ``Frequency-tuned salient region detection,'' in \emph{{Conference on Computer Vision and Pattern Recognition}}.\hskip 1em plus 0.5em minus 0.4em\relax Miami, FL, USA: IEEE, 2009, pp. 1597--1604, {{DOI}: \href{10.1109/CVPR.2009.5206596}{10.1109/CVPR.2009.5206596}}.

\bibitem{Fan2018Enhanced}
D.-P. Fan, C.~Gong, Y.~Cao, B.~Ren, M.-M. Cheng, and A.~Borji, ``Enhanced-alignment measure for binary foreground map evaluation,'' in \emph{{International Joint Conference on Artificial Intelligence}}.\hskip 1em plus 0.5em minus 0.4em\relax Stockholm, Sweden: IJCAI, 2018, pp. 698--704, {{DOI}: \href{10.24963/ijcai.2018/97}{10.24963/ijcai.2018/97}}.

\bibitem{li2023large}
Y.~Li, Q.~Hou, Z.~Zheng, M.-M. Cheng, J.~Yang, and X.~Li, ``Large selective kernel network for remote sensing object detection,'' \emph{arXiv preprint arXiv:2303.09030}, 2023.

\bibitem{molchanov2019importance}
P.~Molchanov, A.~Mallya, S.~Tyree, I.~Frosio, and J.~Kautz, ``Importance estimation for neural network pruning,'' in \emph{Proceedings of the IEEE Conference on Computer Vision and Pattern Recognition}, 2019, pp. 11\,264--11\,272.

\bibitem{gou2021knowledge}
J.~Gou, B.~Yu, S.~J. Maybank, and D.~Tao, ``Knowledge distillation: A survey,'' \emph{International Journal of Computer Vision}, vol. 129, pp. 1789--1819, 2021.

\end{thebibliography}

\end{document}